\documentclass[letterpaper]{article}
\usepackage{aaai}
\usepackage{times}
\usepackage{helvet}
\usepackage{courier}
\usepackage{mathtools}
\usepackage{graphicx}
\usepackage[ruled]{algorithm2e}
\usepackage{amssymb}
\usepackage{changepage}
\usepackage{cleveref}
\usepackage{subfigure}
\crefname{section}{section}{sections}
\crefname{subsection}{subsection}{subsections}
\graphicspath{ {images/} }

\begin{document}
\title{Explore, Exploit or Listen: Combining Human Feedback and Policy Model to Speed up Deep Reinforcement Learning in 3D Worlds}
\author{Zhiyu Lin, Brent Harrison, Aaron Keech, and Mark O. Riedl\\
School of Interactive Computing\\
    Georgia Institute of Technology\\
    Atlanta, GA\\
   \textit{[zhiyulin; aaronkeech; brent.harrison; riedl]@gatech.edu}
}
%Zhiyu Lin, Brent Harrison, Keech Aaron, Mark Riedl
\maketitle

\begin{abstract}
\begin{quote}
We describe a method to use discrete human feedback to enhance the performance of deep learning agents in virtual three-dimensional environments by extending deep-reinforcement learning to model the confidence and consistency of human feedback. 
This enables deep reinforcement learning algorithms to determine the most appropriate time to listen to the human feedback, exploit the current policy model, or explore the agent's environment. 
%We introduce a technique extending deep reinforcement learning that models the confidence and consistency of human feedback and determines when to listen to human feedback, exploit the current policy model, or when to explore. 
Managing the trade-off between these three strategies allows DRL agents to be robust to inconsistent or intermittent human feedback. 
%Making the tradeoff between three strategies is essential because human feedback may not be given for every time step and further may include inconsistent advice.
Through experimentation using a synthetic oracle, we show that our technique improves the training speed and overall performance of deep reinforcement learning in navigating three-dimensional environments using Minecraft.
We further show that our technique is robust to highly innacurate human feedback and can also operate when no human feedback is given.

% whereby we learn human feedbacks to training sessions of such agent using a statistical approach, controlling the chance of the agent utilizing human feedbacks or making decision by itself by a Confidence Check and a Consistency Check. Our method improves the state-of-the-art Deep Reinforcement Learning approach in its training speed and overall performance on multiple environments, even when human feedbacks are inaccurate.
% WIll go back after all the text including experiments and discussions are finished%
\end{quote}
\end{abstract}

\section{Introduction}
%\noindent Machine Learning is a popular and effective method to teach Artificial Intelligence (AI) to do specific tasks. Among different implementations toward this goal, one of them is Interactive Machine Learning (IML) \cite{fails2003interactive}.
\noindent Interactive machine learning (IML)~\cite{fails2003interactive} seeks to improve upon traditional machine learning algorithms by allowing humans to play a direct role in training by providing demonstrations of correct behavior or by actively critiquing the model while the agent is learning. 
This has proven to especially effective at speeding up learning in complex sequential decision making environments that are often solved using reinforcement learning (RL).
This is because the human feedback can be used to enable the agent to explore reasonable behavior trajectories when it would otherwise have no prior knowledge to dictate behavior. 

Human feedback in IML can take different forms. 
{\em Learning from Demonstration} allows humans to directly provide examples of proper behavior\cite{argall2009survey}.
The agent can learn the policy directly, learn to explore more effectively~\cite{NIPS2013_5187}, or learn a reward function from which to reconstruct a policy~\cite{Abbeel:2004:ALV:1015330.1015430}.
It is not always feasible to provide demonstrations.
{\em Learning from Critique} allows human teachers to indicate that the agent is doing well or not doing well in order to bias the agent toward certain outcomes. 
Learning from Critique can also include human indication of preferences over variations in agent behavior~\cite{christiano2017deep}.
{\em Learning from Advice} is similar to Learning from Critique, except the human teacher advises the agent on the actions it should take. 
%It is a proactive form of critique---critique can be turned into action advice by asking humans to indicate their preference for actions.
Preliminary experiments (in preparation) show that humans prefer giving action advice over critique. 
This paper looks at incorporating advice from human teachers into deep reinforcement learning.

Recently, reinforcement learning approaches augmented with deep neural networks have proven to be effective at learning policies in complex sequential environments, such as Atari games~\cite{mnih2013playing} or Minecraft~\cite{johnson2016malmo,oh2016control}, with only access to pixel-level state representations. 
%IML creates an interface for human to interact with AI agents, so that the human is able to provide suggestions in the training process and on the trained classifiers.
While Deep Reinforcement Learning (DRL) is effective at learning control policies in these environments, they are data inefficient in that it can require a large amount of learning episodes and exploration to learn a reasonable policy. 
This often makes these techniques ill-suited on complex, real-world problems. 
To address this limitation, we seek to extend current deep reinforcement learning techniques to enable them to learn from discrete human feedback.
By allowing DRL techniques to learn from human feedback, we seek to drastically reduce the required number of training episodes to learn a reasonable behavior policy in complex virtual worlds. 

%We look into the specific case of training the agent to complete a real-world-style task which a sequence of actions is necessary for the task to be considered completed.
%In this case, suggestions are applied to a local related part of the dataset, which is different from a typical data classifier (like an image recognizer). 
%\subsection{Interactive Machine Learning}

Dealing with human feedback is not a trivial task due to many factors including:
\begin{itemize}
\item \textbf{Inconsistency.} 
It is unlikely that a human teacher will be able to consistently provide correct feedback. 
%This problem only gets worse when we consider the possibility of having multiple teachers training a single agent. 
These inconsistencies can arise because the human trainer does not have a complete grasp of how to complete the task themselves, or possibly because of fatigue or simply making mistakes. 
\item \textbf{Intermittent Feedback.} 
It is also not guaranteed that humans will provide feedback on each action that the agent takes. This means that the human reward signal provided to the agent could be very sparse, further complicating the learning process.
\item \textbf{Differing scales of reward and feedback.} 
In the simplest interpretation, human feedback can be considered part of the reward signal.
However, if there is reward emanating from the environment (a common assumption), then the scale of the environmental reward and scale of feedback values must be tuned to achieve peak learning performance.
\item \textbf{Latent states.}
In DRL in 3D worlds the true state of the agent and environment must be inferred from pixel-level observations which may contain a nontrivial amount of sensor noise. 
Unlike discrete environments, it becomes difficult to determine which states to apply human feedback values.
\end{itemize}

In this paper, we extend off-policy DRL techniques to learn from human advice while taking into account that this advice may be inconsistent and intermittent.
Off-policy reinforcement learners balance exploitation of policy and pure exploration by occasionally selecting a random action. 
Whereas most IML has explored discrete environments---often 2D games---we further show that Learning from Advice can work in 3D worlds where environmental reward and human feedback values are on different scales. 
We evaluate our technique in the three-dimensional virtual world, Minecraft, using a set of simulated oracles meant to mimic human trainers of varying training proficiencies. 

%%%%%%%%%%%%%%%%%%%%%%%%%%%%%%%%%%%%%%%%%%%%%%%%%%%

\section{Preliminaries}

A Markov Decision Process \cite{puterman2014markov} (MDP) is a model used to describe potentially stochastic sequential decision making problem. 

A MDP can be expressed as a tuple $\langle S,A,R,T\rangle$ which contains:
\begin{itemize}
\item A set of possible world states $S$
\item A set of possible agent actions $A$
\item A reward function $R(s,a):S \bigtimes A \rightarrow \mathbb{R}$
\item A transition function $Pr(s'|s, a):S \bigtimes A \bigtimes S' \rightarrow p \in [0,1]$ of each action’s possible effects in each state.
\end{itemize}

The solution to a MDP is a policy $\pi:S \rightarrow a$, which is a function that dictates the best action an agent can take in any world state in order to maximize future rewards. 
Reinforcement Learning is a technique that learns an optimal policy for a MDP by stochastically performing actions and observing their effect on the world. 

%group of approaches to train an AI agent in an environment described as a MDP. 
%The agent is given the goal of collecting as many reward \(\Sigma R\) as possible. 

Q-Learning \cite{watkins1992q} is one of the approaches in RL to help AI agents to approximate a reward function.  An estimate of state-action values, $Q(s,a)$, is iteratively updated in each learning phase as follows:
\begin{equation}
Q(s,a)=Q(s,a)+\alpha[R(s)+\gamma\underset{a'}{\max}\: Q_{old}(s',a')-Q(s,a)]
\end{equation}
where $\gamma$ is a predefined discount factor and $\alpha$ is the learning rate.

For large or unknown environments, Deep Q-Learning \cite{pmlr-v48-gu16} is an extension of original Q-Learning that utilizes a deep neural network to approximate the $Q(s,a)$ even when the number of states $s$ is large.

%%%%%%%%%%%%%%%%%%%%%%%%%%%%%%%%%%%%%%%%%%%%%%%%%%%

\section{Related Work}

There has been much work done on incorporating demonstrations \cite{schaal1997learning,wilson2012bayesian,wirth2016model,akrour2011preference} and critique \cite{argall2009survey,wilson2012bayesian,daniel2015active,wirth2016model,christiano2017deep} into machine learning.
These approaches have proven effective at speeding up learning in complex, sequential environments. 
Typically these methods assume the existence of a reward function and use human feedback to aid the agent in learing a policy that maximizes that reward. 
Inverse reinforcement learning \cite{Abbeel:2004:ALV:1015330.1015430,el2016score}, on the other hand, seeks to directly engineer a reward function based on examples of optimal behavior provided by human trainers. 

% akrour2011: rank-policyonly
% wilson2012: compare2-path-policy only
% daniel2015: rating-policyonly-asreward
% el2016: Inverse ML
% wang2016: Language process + interactive ML
% wirth2016: compare2-path-policy only 
Video games are complex virtual worlds that often emulate many of the complexities found in the real world.
Thus, many machine learning researchers have taken an interest in using machine learning to train AI agents to play video games \cite{mnih2013playing,NIPS2013_5187}.
%is an emerging field of research  due to the ways in which games emulate the complexities of the real world. 
So far, there have been successes in using machine learning in both 2-D \cite{mnih2013playing,mnih2015human} and 3-D environments \cite{NIPS2013_5187,oh2016control,christiano2017deep}. 
%Many works focus on 2-D environments \cite{mnih2013playing,mnih2015human}, others developed state-of-the-art approaches in 3-D environments\cite{NIPS2013_5187,oh2016control,christiano2017deep}. Works on Minecraft utilizing Deep Q-Learning \cite{oh2016control} has yielded positive results, proved its effectiveness in multiple vastly different yet practical environments.

There are studies focusing on combining the reward learning methods with human input, such as \cite{judah2010reinforcement,NIPS2013_5187}. We seek to extend this work in this paper, especially the work performed in \cite{NIPS2013_5187}. Their method aims to rescale the human feedback and generate a universal value.
We extended it by utilizing probabilistic approaches and deep reinforcement techniques to enable similar methods to be adapted to continuous state space.
% - Policy Shaping

A noticeable difference between our study with others is that ours are based on Deep Q-Learning. Instead of a discretized state space, our assumption is that the state space can be continuous thus a finite array of states can not be easily defined. Since Deep Q-Learning is a relatively new topic compared to the long history of Learning from Demonstration, many techniques are based on assumptions that no longer stand in 3-D environments and we are unable to apply their methods directly nor compare our method with theirs. 
%However, many ideas in the former research \cite{mnih2013playing,NIPS2013_5187} do inspire our study.

In the context of incorporating human feedback into a deep reinforcement learning paradigm, Christiano et al. \shortcite{christiano2017deep} collects preferences over action trajectories and trains a neural network to produce reward values. 
%In this work, human feedbacks are collected as a result of comparison between 2 trajectories similar to \cite{wilson2012bayesian}. 
This is similar to the work by Wilson, Fern and Tadapalli \shortcite{wilson2012bayesian} which uses preferences over trajectories in Bayesian policy learning.
Though the human feedback gathering interface of our methods and theirs share similar traits, our method has fundamental differences with theirs. 
Our work differs from these approaches in that we combine environmental reward and human feedback instead of relying only on preference feedback.
%Since our goal is to improve typical Q-Learning by embedding human feedback, we safely assume that a precise reward can be achieved since this is the prerequisite for typical Deep Q-Learning to work, and 
Video games have clear rewards and the goal of our work is to augment agent ability to learn a policy in environments where it is difficult to learn an optimal policy even with the presence environmental rewards. 
%Their work is based on the exact opposite that reward functions are not available, thus typical Q-Learning will not work and is not used. However, 
We believe that our method and theirs can be combined to achieve better performance and reduce the feedback query frequency, but we leave it to future work.

\section{Method}
\newcommand\defineAs{\stackrel{\mathclap{\normalfont\mbox{def}}}{=}}

% \label{something}, \ref{something} to reference back
% \label, \eqref
%\subsection{Overview}

The challenges of incorporating human feedback into deep reinforcement learning are two-fold.
First, we must construct a deep neural network that learns to map pixels from an agent's sensors into $Q$-values---called a Deep Q Network (DQN). 
Much of prior work on interactive machine learning that incorporates human critique feedback \cite{NIPS2013_5187} was performed in discrete environments with no sensor error.
In these environments, states can be uniquely matched when determining how to adjust the $Q$-values of states based on human feedback.

Second, if there is a reward signal from the environment then scaling the human feedback and/or environment reward appropriately is a non-trivial tuning problem.
An environmental reward signal that is strong relative to the human feedback may make an agent incapable of learning from human feedback. 
This is a well-known problem in {\em modular reinforcement learning} \cite{simpkins2010integrating}.
A large amount of feedback may also overwhelm environmental rewards.
Further complicating the situation, a human trainer may not choose to provide feedback at every state.
It is thus also non-trivial how to handle ``silence''. %\cite{cederborg}. 
However, even the naive solution of rendering lack of feedback as a zero can be problematic when the environmental reward scale allows for negative reward values.
%
%The work by Christiano et al. \shortcite{christiano2017deep} avoids these challenges by not having any environmental reward and relying solely on a reward function learned from human indications of preferences.

\begin{figure}
\centering
\includegraphics[width=3.5in]{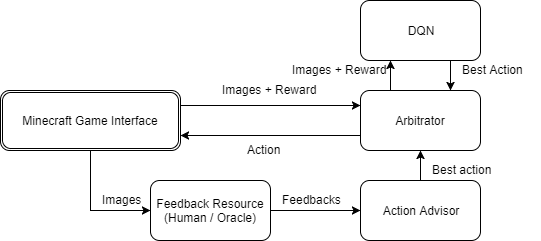}
\caption{The agent architecture.}
\label{fig:structure}
\end{figure}

In this section, we describe a technique for incorporating human feedback into a deep reinforcement learning paradigm.
Our technique makes use of an {\em arbiter}, which decides when to execute an action computed by a deep Q network or to follow human action advice (Figure \ref{fig:structure}).
The arbiter measures the confidence in the deep Q network as it learns---a function of network loss---and the consensus between the human action advice and the action selected by the deep Q network.
The arbiter then picks between random exploration, exploiting the action picked by the deep Q network, and exploiting the human action advice (if any).
Our technique is thus an off-policy reinforcement learner.

Figure~\ref{fig:structure} shows the system architecture for our technique.
The Minecraft environment and deep Q network modules form the standard reinforcement learning loop---the environment produces a state and a reward, which is consumed by the learner and an action is executed in the environment. 
In the case of 3D virtual environments such as Minecraft, the state is an image generated from the first-person perspective of the situated agent.
A second loop exists where a human oracle (or synthetic oracle for experimentation purposes) is also observing the environment state and {\em may} choose to provide feedback in the form of action advice.
Action advice indicates which action(s) the oracle believes the agent should take.
As noted above, the arbiter sits in the middle of both loops and must choose between the action computed by the DQN and the action advice of the oracle.

%In this section, we describe the approach we utilize information from multiple learners to boost training performance. The overall structure and data flow of the implementation of our method is presented in \cref{fig:structure}. 

%We allow all the learners to present their action suggestions independently with a confidence index, then use a scheduler to decide how we treat with them and combine them into a single action as the next step in the training process.

\subsection{Deep Q Network}
% We define a \textbf{Generalized Learner} as a data source which when being fed by environment information, will give a best action suggestion. 
% \begin{figure}
% \caption{A typical learner.}
% \centering
% \includegraphics[width=3.5in]{learner}
% \end{figure}

%The two learners we used in our method is as follows:

The Deep Q Network (DQN) is a neural network to approximate $Q$ values for states---in this case first person perspective images.
The DQN returns the action with the highest predicted $Q$ value to the arbiter. 
We implemented the DQN used within Mnih et al.~\shortcite{mnih2013playing}, adapted for images from the 3D environment and for the Minecraft controls.
Our agent is an off-policy learner because the DQN produces the action with the best predicted $Q$ value based on network parameters and then the arbiter (described below) determines whether to execute the action or pick a random action instead. 
%as the actions the agent takes are generated by combining network parameters as well as this arbiter module.

%Deep Q-Learner, which is a deep learner utilizing Q-Learning and Convolutional Neural Network\cite{krizhevsky2012imagenet}. The original research\cite{mnih2013playing} are targeted at a group of 2-Dimensional Atari games; We re-purposed it for our 3-Dimensional environment with different input and output layers, and used it as our baseline.

%In all the experiments we used hand-tuned parameters to maximize performance. 
%The Deep Q Network technique by Mnih et al. uses an epsilon-greedy exploration/exploitation strategy. We decay $\epsilon$ according to the following schedule:
%We used $t_{min}$ = 600 and $t_{max} = 2000$ in the epsilon-greedy process in the baseline (using reward learner \cite{mnih2013playing} only.) The decay function is as follows:

\subsection{Action Advisor}

The Action Advisor acts as an interface between a human (or synthetic) oracle.
The Action Advisor queries the oracle for each possible action, asking if it would be a good or bad action to take in the currently observed state.
The reason we ask the oracle for a binary decision for each possible action is because there may be multiple acceptable actions. 
The Action Advisor uniformly selects a single action from the actions indicated to be good and passes it to the arbiter.

Note that it would be trivial to extend the Action Advisor to receive a non-binary preference for each action and sample an action from a weighted distribution over action preferences.

%Policy Learner, which serves as an interface to present feedbacks into the system. The policy learner queries information from out-of-system information source and generates a best action based on the information gathered. Each query contains an image (state) and all possible actions in that state and asks for \textbf{Whether taking one of the action in such image state is "Good" or "Bad".} We call them "good feedback" and "bad feedback" respectively. We then select the action corresponding to the "good feedback". We assume that only one action will be given "good feedback" thus the consequences of providing multiple "good feedback" in the same image state is undetermined; However, in our experimental implementation we take the first "good feedback" provided for the same image state, or a default action if no "good feedback" is observed.

\subsection{Arbiter: Aggregating Action Suggestions}

In Deep Reinforcement Learning, aggregating information from multiple sources is not a trivial task. Risks exist in different point of transferring human feedbacks into the system, including the risk of human errors, inconsistency and limited exploration leading to incomplete learning space. We mitigate these risks using a probabilistic approach wherein ``checks'' are used decide the exploration schedule, or whether the agent should depend on its deep Q network, or listen to human oracle.
The checks used in our technique consist of an: exploration check, confidence check, and consensus check.

%We achieved this by using a \textbf{Learning Scheduler}. Our Learning Scheduler is based on two assumptions based on human instinct:
% \begin{itemize}
% \item \textbf{Stubborn Instinct}, "When I have confidence in my decision, I'll believe in it and follow it more than hearing from others."
% \item \textbf{Consensus Instinct}, "When people agree with me, I will tend to think it is correct."
% \end{itemize}

%\begin{itemize}
%\item 
\subsubsection{Exploration Check}
%\textbf{Exploration Check}, which if passed, 
If passed, this check
guides the agent to do a random action; neither the DQN nor the Action Advisor is queried.
The DQN without the Arbiter implements a typical off-policy exploitation strategy based on past experience. %when the Q-network is highly uncertain.
However, the exploration check makes the agent follow an epsilon-greedy strategy, forcing the agent to do exploration some percentage of the time.
%do more exploration of the environment early during training, which has the benefit of providing a lot of unique observations to the deep Q network.
The likelihood of passing the exploration check decays over time.
%This check is to give agent chance to explore the field randomly in the beginning of the training session, and to allow new states to be observable even in the later part of the session;
The probability of passing the exploration check is computed as:

\begin{equation}
p_{\mathrm{explore}} = \begin{cases} 
      1 & t < r_{min} \\
      e^{\ln{0.01}*\frac{t-t_{min}}{t_{max}-t_{min}}} & t_{min}\leq t < t_{max} \\
      0.01 & t_{max}\leq t 
   \end{cases}
\end{equation}

\noindent
In experiments, we used \(r_{min} = 600\) and \(r_{max} = 2000 \) in our experiments. 

%\item 
\subsubsection{Confidence Check}
%\textbf{Confidence Check}, which 
This check
uses a measure on how confident the agent is 
in the suggestion of the DQN.
When the confidence in the DQN is low, the Arbiter will prefer the action suggestion (if any) from the Action Advisor.
When the confidence in the DQN is high, the Arbiter reduces the frequency that it requests advice from the oracle.
%confident on its decision. Mirroring the human activities when a better confidence level led to a more stubborn instinct to not hear from others, this check allows the agent to learn from human more often in the beginning of the training session, and reduce the frequency of requesting human feedbacks in the later stage base on the demand;
The probability of passing the confidence check is computed as:

\begin{equation}
p_{\mathrm{conf}} = 
\frac{-1}{\ln{\sqrt{\frac{l}{l_{\mathrm{max}}}}}-1}
\end{equation}
where $l$ is the loss of the DQN and $l_{\mathrm{max}}$ is the highest loss observed so far.

%\item 
\subsubsection{Consensus Check}
%\textbf{Consensus Check}, which 
This check
uses a measure of how the best action from the DQN aligns with action advice from the oracle. 
The consensus check is used to counteract inconsistency in the oracle---human oracles are known to be noisy and their ability to consistently provide correct feedback can vary.
When the oracle is inconsistent, and thus there is no consensus with the DQN from time interval to time interval, the arbiter relies more on the DQN.
%This counter hostile human feedbacks so that the agent can auto-correct itself when human feedbacks are poor in performance or even invalid.
The probability of passing the consensus check is a function of the consensus probability at previous time steps:

\begin{equation}
p_{\mathrm{cons},t} = \begin{cases} 
      \max({1,p_{\mathrm{cons},t-1})*f_{1}*d} & a_{\mathrm{DQN}} = a_{\mathrm{AA}} \\
      p_{\mathrm{cons},t-1}*f_{2}*d & a_{\mathrm{DQN}} \neq a_{\mathrm{AA}}
   \end{cases}
\end{equation}
where
$a_{\mathrm{DQN}}$ is the action suggestion from the DQN and $a_{\mathrm{AA}}$ is the action suggestion from the Action Advisor, and
\begin{equation}
d = \begin{cases}1.001 & p_{\mathrm{cons},t-1} < 0.5 \\ 0.999 & p_{\mathrm{cons},t-1} > 0.5\end{cases}
\end{equation}
and $f_{1} = 1.004$ and $f_{2} = 0.998$.
The values of $f_1$ and $f_2$ were found to empirically work well, though any value greater than 1.0 for $f_1$ and any value less than 1.0 for $f_2$ should work. 

%\end{itemize}

%\noindent
%The arbiter algorithm is shown in Algorithm~\ref{alg:arbiter}.

%The implementation of these checks are based on the confidence score from the learners so are learner-specific. See section \ref{section:model-parameters} for implementation details.

%Although in this paper we described only two sources of action suggestions, the technique is easily extensible to three or more sources of action suggestions.

\begin{algorithm}[t]
\caption{The arbiter algorithm used during agent training.}
\label{alg:arbiter}
\SetKwIF{With}{ElseWith}{Else}{with}{do}{else with}{else}{end}
\SetAlgoLined
Initialize all learners and training parameters\;
\While{training goal not achieved\;}{
%Get suggestions from reward and policy learners\;
Calculate $p_{\mathrm{explore}}$, $p_{\mathrm{conf}}$ and $p_{\mathrm{cons}}$\;
\With{probability $p_{\mathrm{explore}}$}
  {\Return{a random action from possible actions\;}}
\ElseWith{probability $p_{\mathrm{conf}}$ {\bf and with} probability $p_{\mathrm{cons}}$ }
  {Request action suggestion from Action Advisor\;
    \Return{action from Action Advisor\;}
  }

\Else{
  {Get action suggestion from DRL\;
   \Return{action from DRL\;}}

}
  }
\SetAlgoLined
\end{algorithm}

\section{Experimental Setup}

%\subsection{Minecraft}

%Minecraft "has been one of the most unusual success stories in gaming in recent memory". \cite{Duncan:2011:MBC:2207096.2207097} 
Minecraft provides players with an open world that they can explore and forge by utilizing a set group of tools to manipulate ``blocks'', which is the main building unit of the game.
The agent is given the task of finding and picking up an object in different maze-like environments.
Navigating the 3-dimensional landscape of Minecraft is an essential part of all tasks the player must perform in the game. 
The task of navigation is also easy to control in terms of difficulty for experimentation purposes and the availability of the game makes it easy to share results.

While the environment appears simpler than other 3D games, such as Doom, because of the use of blocks, the visual simplicity results in instances of ``aliasing'' where spaces can be nearly indistinguishable.
%\todo{do we want to talk about aliasing, or has that become less important?}
Human feedback is especially important in situations of aliasing because the human teacher often has an intuitive understanding that different areas that appear similar may not be treated identically.
With regard to other 3D environments such as Doom, Minecraft also makes much more use of non-enclosed spaces.

We have developed a Minecraft clone, implemented in Unity3D but sharing aesthetic and functional similarity with the original game. 
This was done to make agent integration and experimentation easier.
%built a simplified Minecraft clone, mainly focused on explore part of the original gameplay. In our clone, the agent is given tasks to find and pick up an object in different environments.
To further control for the complexity of the environment, the agent is only allowed to face four 90-degree separated directions (North, South, East, West); However, we perturb the viewing angle randomly after each action by rotating the viewing angle by up to 2 degrees to the right or the left.
%the there can be a slight viewpoint change even when the same state is entered, making every play unique. 

\begin{figure}
\includegraphics[width=3.3in]{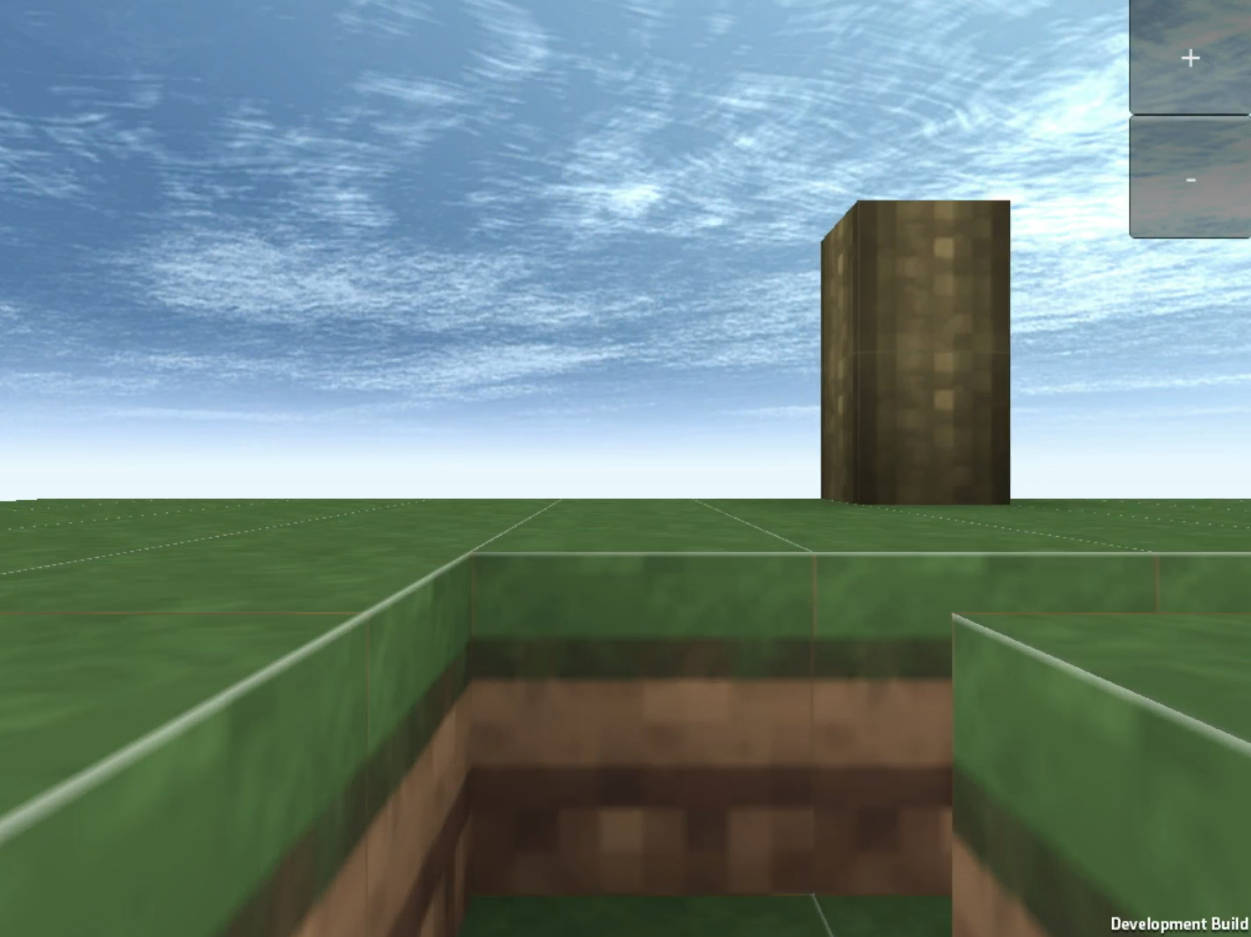}
\caption{A sample screenshot of the agent playing.}
\label{fig:screenshot}
\end{figure}

\begin{figure*}[tb!]
\centering
\subfigure[Easy map]{
	\includegraphics[width=2.5in]{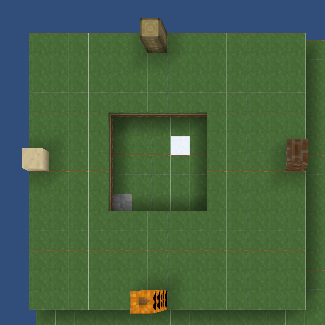}
	\label{fig:map-easy}
}
\subfigure[Hard map]{
	\includegraphics[width=2.5in]{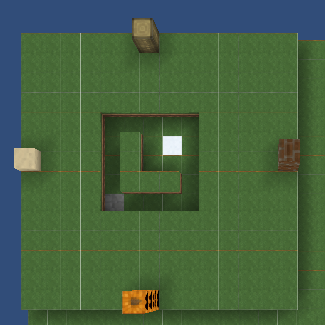}
	\label{fig:map-hard}
}

% \begin{subfigure}{0.49\linewidth}
% \includegraphics[width=0.7\linewidth]{map-hard}
% \caption{"Hard" Map}
% \end{subfigure}
% ~
% \begin{subfigure}{0.49\linewidth}
% \includegraphics[width=0.7\linewidth]{map-easy}
% \caption{"Easy" Map}
% \end{subfigure}
\caption{Top-down view of ``easy'' and ``hard'' maps used in our experiments. The blue block is the start and the quite block is the location of the object to be picked up. Pillars are used to substitute for landscape details that partially disambiguate first-person views.}
\label{fig:env}
\end{figure*}

Since training in graphical 3D worlds creates a large computation and time overhead, we built a state cache system so that we don't have to run the graphical client for every trial. 
%to provide offline training and to decrease the overhead to a minimum level. 
For each state---each block the agent can stand on for and each of the four cardinal directions the agent can face---we store a number of screenshots corresponding to different perturbations.
The Minecraft clone tracks the true state of the agent and randomly selects an image to pass to the reinforcement learning agent.

%\subsection{Environment Design}

We designed multiple different tasks at differing levels of difficulty (Figure~\ref{fig:screenshot}. 
%in regard of how complex they are. 
By identifying a task as a complex one, we looked into the opportunity cost of a non-optimal action. A task that is more complex than another one in this sense has more and deeper dead ends and less freedom of error correction.

Agents are given rewards on actions taken: They get -1 for each step taken and 100 for picking up the object and thus completing the task.

%\subsection{Constructing an Oracle}

In our experiments, we used simulated oracle in the place of a human teacher. 
This is a standard technique (cf., \cite{NIPS2013_5187}) that grants us abilities to systematically manipulate feedback accuracy and test different parameters of environment and hypothetical teacher. 
The oracle was created by labeling each actual agent state (position, orientation) with the best action.
Since human teachers can be inaccurate and have different rates of response to requests for feedback, the simulated oracle allows us to parameterize the accuracy and frequency of oracle response by randomly choosing one of possible feedback outcomes.
%(``Good'' or ``Bad'').

\section{Experiments}

%\subsection {Our metrics}

Due to the nature of Deep Q-Learning, the learned policy can be very noisy~\cite{mnih2013playing}. However, we decided to keep that trait, using a moving average of performance since it applies to an infinite and/or loop-containing non-terminal policy, which if $Q$-value sums are used can lead to confusion in evaluations.
After each training session is over---by finishing the task or timeout---we evaluate the agent by asking it to do the task by itself (e.g., run the policy)  then record its final reward as performance.

In our experiments we compare our method with a baseline, an ablated version of our technique with the Action Advisor removed.
This baseline agent is equivalent to the deep reinforcement learning technique introduced by Mnih et al. \cite{mnih2013playing} with adaptations of the input and output layers of the neural network to Minecraft. 
The baseline additionally uses a different epsilon decay schedule, which we empirically tuned to strengthen the baseline:

\begin{equation}
\epsilon = \begin{cases} 
      1 & t < r_{min} \\
      1-0.9999*\frac{t-t_{min}}{t_{max}-t_{min}} & t_{min}\leq t < t_{max} \\
      0.0001 & t_{max}\leq t 
   \end{cases}
\end{equation}

\noindent
where $t$ is the training episode and $t_{min}$ = 600 and $t_{max} = 2000$.

\subsection {Performance with Consistent and Accurate Oracle}

\begin{figure*}[tb]
\centering
\subfigure[Performance on the hard map]{
	\includegraphics[width=3.3in]{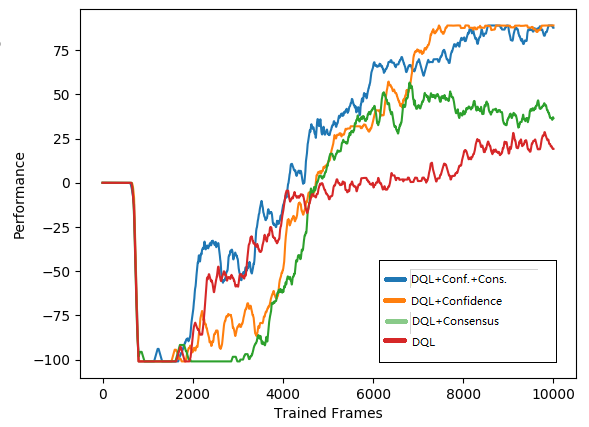}
    \label{fig:perf-hard}
}
\subfigure[Performance on the easy map]{
	\includegraphics[width=3.3in]{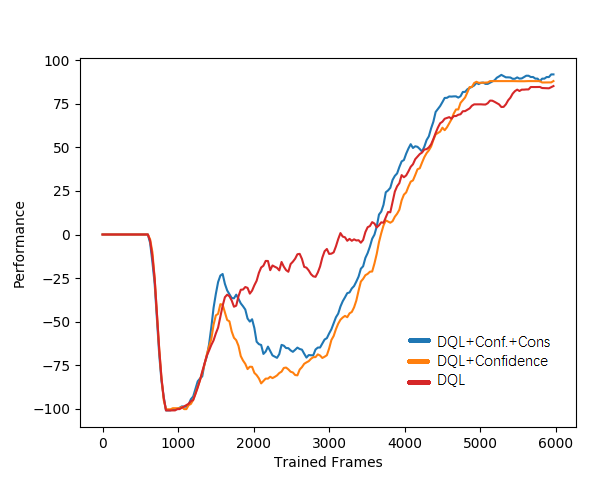}
    \label{fig:perf-easy}
}
% \begin{subfigure}{0.49\linewidth}
% \includegraphics[width=0.7\linewidth]{perf-easy}
% \caption{"Hard" Map}
% \end{subfigure}
% ~
% \begin{subfigure}{0.49\linewidth}
% \includegraphics[width=0.7\linewidth]{perf-easy}
% \caption{"Easy" map}
% \end{subfigure}

\caption{Performance of our method comparing to the baseline (DQN only). Each entry represents the average performance from 20 training sessions.}
\label{fig:performance_perfect}
\end{figure*}

\begin{figure*}[tb!]
\centering
\subfigure[Baseline]{
	\includegraphics[width=2.1in]{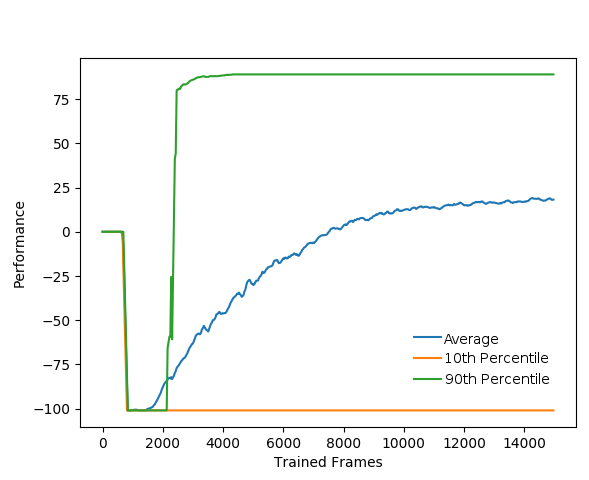}
}
\subfigure[Condifence check only]{
	\includegraphics[width=2.1in]{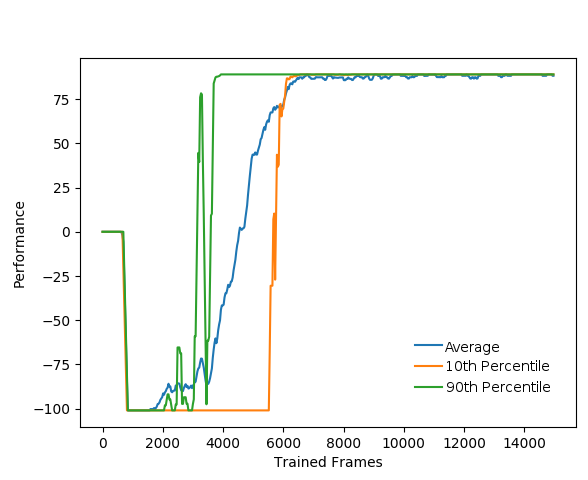}
}
\subfigure[Confidence and Consistency Check]{
	\includegraphics[width=2.1in]{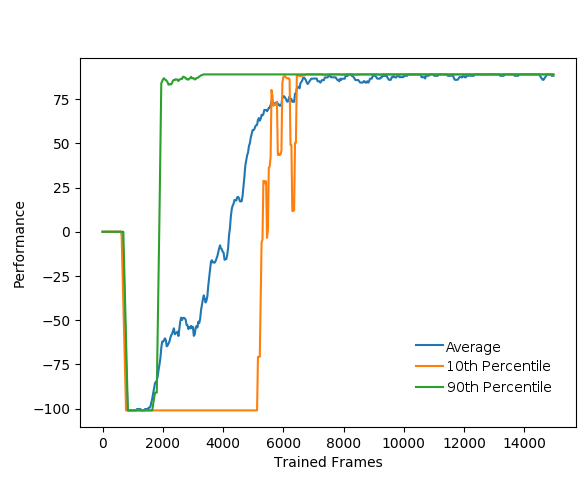}
}
%\subfigure[]{
%	\includegraphics[width=1in]{legend-performance}
%}

% % \begin{subfigure}{0.11\linewidth}
% % \includegraphics[width=.99\linewidth]{legend-performance}
% % \end{subfigure}
\caption{Performance comparison on the "hard" map in detail.}
\label{fig:all_comparison}
\end{figure*}

In the ideal case of consistent and accurate oracle feedback, Figure~\ref{fig:performance_perfect} shows that our technique converges on the optimal policy faster than the baseline. 
We evaluated our method on the two maps from Figure~\ref{fig:env}.
Our results show that our technique never under-performs the baseline.
The agent doesn't benefit from oracle feedback in the easy map because it is relatively trivial to find the optimal policy on the easy map.
Two maps are sufficient to show the trend: as the map becomes more maze-like and the number of obstacles increases, the importance successfully incorporating oracle feedback into policy learning is evident.

In the case of consistent and accurate oracle feedback, the consensus check slightly reduces learning performance. 
Oracle feedback is considered by the Arbiter via the confidence check---if confidence in the DQN is low, seek advice from the oracle---and through the consensus check.
Disabling the confidence check or consensus check affects how the agent uses the oracle feedback. 
The confidence check makes the most effective use of oracle feedback.
This makes sense because the confidence check is guiding the agent toward the most reliable action suggestion.
The consensus check, on the other hand, warns the agent away from malicious oracle action advice.

% \begin{itemize}
% \item More complex environment observes better performance boost in training speed;
% \item Confidence check expressing the most performance boost;
% \item Combining two checks have little to no benefit.
% \end{itemize}

Figure~\ref{fig:all_comparison} shows the reward as training increases for the baseline, our technique with confidence check, and our technique with both confidence and consensus checks. 
The figure shows the average reward, the 90th percentile (90\% of all trials perform no better than this), and the 10th percentile (10\% of all trials perform no better than this).
The charts show that there is a very wide variance in performance for the baseline. 
Sometimes the learner gets lucky and hits upon a good policy, but often it times out before finding an acceptable policy.
Our technique with just the confidence check or with confidence and consensus checks significantly reduces the variance in learning performance, making it much more reliably able to find the optimal policy.

The decreased average for the consensus-only agent (Figure~\ref{fig:perf-hard}) is due to a slightly increased variance---some small percentage of the time the agent will time out during trials resulting in a very low reward value being averaged in.

\begin{figure}[tb]
\centering
\includegraphics[width=2.5in]{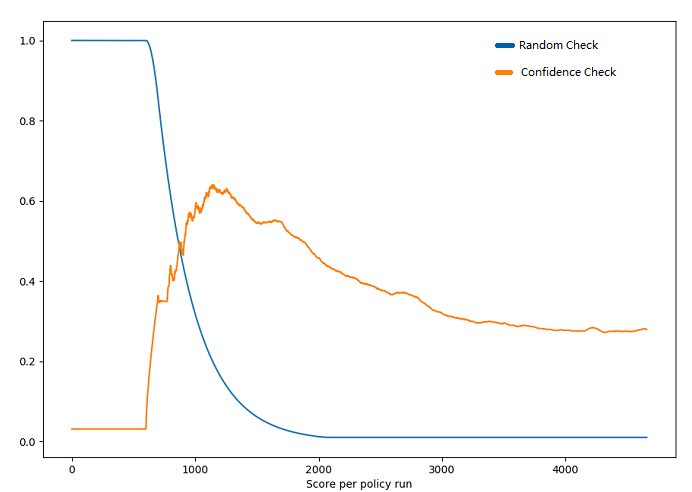}
\caption{Probability of Random and Confidence Check passing in a typical run on "Hard" map. Trace ends at convergence.}
\label{fig:schedule-prob}
\end{figure}

We also inspect the effect of our technique on requests for feedback from the oracle.
Figure~\ref{fig:schedule-prob} shows the probability that the agent requests feedback from the oracle over time as training progresses.
The function for random exploration (epsilon decay) is shown for reference.
Note that the bulk of feedback requests come after an initial period of intense exploration and then decreases as the confidence in the DQN increases.

\subsection {Performance with Differing Oracle Accuracy}

\begin{figure*}[tb!]
\centering
\subfigure[Confidence Check Only]{
	\includegraphics[width=3.3in]{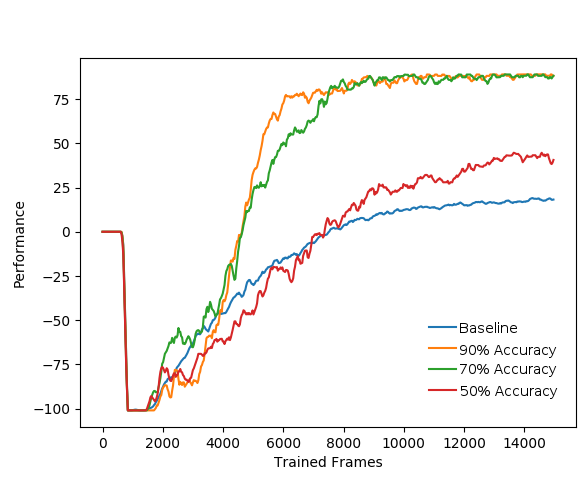}
}
\subfigure[Confidence and Consensus Check]{
	\includegraphics[width=3.3in]{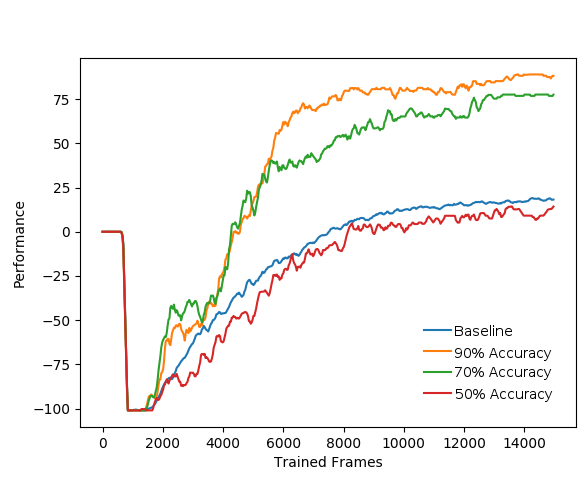}
}
% \begin{subfigure}{0.4\linewidth}
% \includegraphics[width=.999\linewidth]{perf-comp-1c}
% \caption{Confidence Check Only}
% \end{subfigure}
% ~
% \begin{subfigure}{0.4\linewidth}
% \includegraphics[width=.999\linewidth]{perf-comp-2c}
% \caption{Confidence and Consistent Check}
% \end{subfigure}
% ~
% \begin{subfigure}{0.15\linewidth}
% \includegraphics[width=.999\linewidth]{legend-accuracy-comp}
% \end{subfigure}
\caption{Performance of our method with varied feedback accuracy on different methods on "hard" map.}
\label{fig:accuracy_comparison}
\end{figure*}

We evaluated our method in non-perfect case, in which we degrade the oracle's accuracy. 
Figure~\ref{fig:accuracy_comparison} shows that the agent is robust against oracle inaccuracy. 
Even at 50\% oracle accuracy (i.e., random feedback) our technique achieves a performance that is equivalent to the baseline. 
Christiano et al. \shortcite{christiano2017deep} assume a 10 percent of human error, well within the scope of our technique's abilities. 

The extent to whether the consensus check helps is unclear.
The consensus check seems to nullify the benefit of feedback as the oracle accuracy decreases.
However, the average for the confidence+consistency check version is due to training timeouts and only a few timeouts can cause the average to plummet. 
Our synthetic non-perfect oracle naively injects random advice into the agent that can be implausible, such as repeatedly advising the agent walk straight into a wall.
Thus we believe that the synthetic oracle can be quite antagonistic at times, deflating the agent performance curves.

This experiment also gives an indication of how the system will respond to ``silence'', since lack of feedback can be interpreted as a ``wrong feedback''.

%\subsection {Performance on different environments}

% \todo{I don't understand how this is different than the first experiment on hard and easy. Are you saying that there are other maps besides the two? Or are we into discussion now? - First experiment is on hard only, this paragraph is talking about hard vs. easy specifically speaking. MOVED THIS EARLIER}
% We evaluated our method on different environments where some obstacles are added or removed compared to the typical map. Our results from Figure~\ref{fig:env} show that our method works on par with or better than the baseline, with the more complex task seeing better performance boost.

%\subsection {How each check contribute to the scene}

% \todo{do we have anything showing this? The figures we have don't do that. ~\ref{fig:performance_perfect}}
% Confidence check clearly contributes most to the performance boost, while Consistency check with itself seems to contribute less. Whether Confidence and Consistency Check combined grants better performance than Confidence check only is uncertain; Our experiments suggests an on-par performance but this can be heavily environment-dependent.
%its effect emerges in imperfect cases. Assuming a typical scene when human are in the process, since people make mistakes, having the Consistency check helps on improving the robustness of the system.

% \begin{figure}
% \caption{Performance with certain learners or checks disabled.}
% \includegraphics{placeholder}
% \end{figure}

\section{Conclusions}

Interactive Machine Learning (IML) postulates that for environments and tasks that present substantial challenges to reinforcement learning algorithms, human teachers can help agents learn.
Learning from Advice---the agent asks a human what they would do in a particular situation---has the potential to significantly improve machine learning outcomes while keeping human feedback overhead manageable.
We have demonstrated that our technique for incorporating human advice in 3D graphical environments performs no worse than non-interactive learners and benefits from advice as the task becomes harder.

Two important properties of our technique are that (a) the human teacher is never required to give feedback to the agent, and (b) the human teacher is not assumed to be perfect.
The former is essential for IML because providing advice has a cognitive burden on the human teacher and it may not be practical for the human teacher to weigh in on every move the agent tries while learning to perform a task.
The latter is essential because humans make mistakes or can become confused themselves.
We show that our technique is robust against human teacher error; as long as the human is not intentionally adversarial (i.e., greater than 50\% accuracy), our agent learning technique can glean some advantage out of human feedback.

From the perspective of designing reinforcement learning agents that incorporate human feedback, our technique presents a general strategy for dealing with the fact that environmental rewards can be arbitrary and that environmental rewards and human feedback can be on drastically different scales. 
The arbiter is only required to decide to exploit the action suggestion of the deep Q network, exploit the action advice from the oracle, or explore the environment with a random action.
It learns a schedule of exploitation, exploration, and listening to advice based on DQN learning rate and the consistency of the human oracle.

% {Mark needs to review}
% This paper defined the paradigm for integrating feedback with Deep Reinforcement Learning.
% We introduced Arbitor, which improves Deep Reinforcement Learning by carefully introducing human feedbacks in the training process. Policy Combiner produced results on par with or better than the Deep Reinforcement Learning techniques alone, with harder tasks having more improvements, and it demonstrated robustness to common level inaccurate feedback. With these
% advancements this paper may help in making combining these two state-of-the-art methods an increasing viable
% option for intelligent systems.

\section*{Acknowledgments}
This material is based upon work supported by the Office of Naval Research (ONR) under Grant \#N00014-14-1-0003.
% Anonymized for blind review.

%\cite{xx}

\bibliography{main}
\bibliographystyle{aaai}
\end{document}